\documentclass{article}

\usepackage{PRIMEarxiv}

\usepackage[utf8]{inputenc} % allow utf-8 input
\usepackage[T1]{fontenc}    % use 8-bit T1 fonts
\usepackage{hyperref}       % hyperlinks
\usepackage{url}            % simple URL typesetting
\usepackage{booktabs}       % professional-quality tables
\usepackage{amsfonts}       % blackboard math symbols
\usepackage{nicefrac}       % compact symbols for 1/2, etc.
\usepackage{microtype}      % microtypography
\usepackage{lipsum}
\usepackage{fancyhdr}       % header
\usepackage{graphicx}       % graphics
\usepackage{authblk}
\usepackage{amsmath}
\usepackage{placeins}
\usepackage{float}
\graphicspath{{media/}}     % organize your images and other figures under media/ folder

\title{ReMAP: Self-supervised learning to unveil brain representations and vulnerability}
\author[1,3,4,5,*]{Jade Perdereau}
\author[1,3,5]{Virginie Loison}
\author[1,3]{Kanssa El Ayeb}
\author[2,3]{Louis Gervais}
\author[1,3]{Melvin Berto Strouc}
\author[1,3,4,5]{Fabrice Vallée}
\author[5]{Thomas Moreau}
\author[1,3,4]{Jérôme Cartailler}

\affil[1]{AP-HP, Hôpital Lariboisière, Paris, France}
\affil[2]{Sorbonne Université, Paris, France}
\affil[3]{UMR-942, Inserm Délégation Régionale Paris 7, Bagnolet, France}
\affil[4]{Université Paris Cité, Boulogne-Billancourt, France}
\affil[5]{Université Paris-Saclay, Inria, CEA, Palaiseau, France}
\affil[*]{Corresponding author: \texttt{jade.perdereau@inria.fr}}

\begin{document}
\maketitle

\begin{abstract}
General anesthesia offers a rare opportunity to observe the human brain under a standardized, controlled perturbation. Yet intraoperative electroencephalography (EEG) is almost always reduced to a single proprietary depth index, collapsing a rich trajectory into one number and discarding how a brain moves between states. Here we ask whether the geometry of that trajectory, not merely the depth it reaches, carries clinically meaningful information. Using similarity-based self-supervised learning on raw, two-electrode frontal EEG, with no labels, we place each recording within a low-dimensional space in which anesthetic depth becomes one readable axis while the shape of a patient's path encodes additional structure. We validate the representation across two cohorts and two acquisition systems totaling more than 1,000 patients. Depth of anesthesia is predicted accurately (BIS mean absolute error = 3.2, R² = 0.82), and in the sparse-montage setting our compact (~68k parameter) model remains competitive with EEG foundation models orders of magnitude larger (4M-157M parameters), indicating that matching the representation to the recording dominates raw scale. The learned space organizes age along its own gradient, independent from depth, without supervision. The same space also aligns with interpretable anesthetic signatures like frontal alpha, slow-delta, and burst suppression, linking this data-driven representation to established neurophysiology. On an independent cohort with longitudinal follow-up, the geometry of the early trajectory separates 30-month cognitive and mortality outcomes complementary to age (AUROC 0.86). These results suggest that the path a brain traces through anesthesia is a label-efficient correlate of latent vulnerability, motivating prospective validation.
\end{abstract}

% keywords can be removed

\section{Introduction}
	 Brain monitoring using simplified frontal electroencephalography (EEG) during general anesthesia is becoming increasingly routine \cite{purdon_clinical_2015, bruhn_depth_2006, klein_recommendations_2021}, providing a low-burden and scalable window onto a patient's brain function during surgery. Despite the richness of the EEG signal, in practice, it is almost always reduced to a single proprietary depth index - such as the Bispectral Index (BIS) or Patient State Index (PSI). This number usefully guides anesthetic titration but discards most of the information the EEG carries. The limitation is dimensional, as a single value reports the depth of anesthesia level well but not the trajectory the brain takes to reach it. A natural alternative is geometric. Rather than summarizing the anaesthetized brain as a single point, one could place it within a space, in which the trajectory of a patient's anesthesia can be interpreted by physicians.

Why this richer representation matters clinically is linked to the role of brain monitoring. Too little anesthetic can lead to accidental awareness with lasting psychological sequelae \cite{ghoneim_incidence_2007, sandin_awareness_2000}. Conversely, intraoperative EEG patterns such as prolonged burst suppression and reduced anesthetic-induced alpha power have been associated with postoperative delirium, cognitive decline, and mortality, and have been interpreted as markers of an underlying systemic vulnerability rather than as modifiable intraoperative causes \cite{fritz_preoperative_2020, dustin_boone_processed_2022, mather_intraoperative_2025}. From this perspective, EEG monitoring should not only be reduced to an anesthesia index, but also leveraged to observe and track how an individual's brain responds when exposed to anesthetic drugs or surgical stress, because that reaction may reveal clinically relevant vulnerability or resilience \cite{purdon_electroencephalogram_2013, brown_general_2010}. Current monitoring captures little of this as depth of anesthesia indices aim at distinguishing mostly wakefulness, normal and deep sedation, but lack granularity, which is why commercial depth indices disagree for identical EEG in roughly two-thirds of cases \cite{hight_five_2023}, and even BIS, the most established of them, has shown inconsistent reduction of intraoperative awareness in high-risk patients \cite{avidan_prevention_2011, avidan_anesthesia_2008, myles_bispectral_2004}.

Learning a richer representation from raw EEG is not fully solved by existing tools. Traditional methods based on hand-engineered spectral features carry direct physiological meaning \cite{xu_intraoperative_2025, akeju_effects_2014} but capture only a narrow slice of the signal \cite{aubin_repurposing_2023, cartailler_brain_2021}. Supervised models hold great potential for clinical applications \cite{chambon_deep_2018} but require a large amount of labelled data, often at an impractical scale \cite{ahn_development_2025}. Novel approaches based on EEG foundation models claim to solve a variety of cross-domain tasks using a unified model trained on large, heterogeneous and multi-channel datasets \cite{wang_cbramod_2025, ouahidi_reve_2025, kostas_bendr_2021}. Yet they do not reliably outperform domain-specific approaches, particularly in sparse montages, clinical and distributional shifts settings \cite{kastrati_eeg-bench_2025, xiong_eeg-fm-bench_2026}. Where intraoperative EEG has been useful on long-term prognosis or vulnerability, it has typically done so using summary features from a short and stable maintenance window in a static model \cite{mather_intraoperative_2025, fritz_preoperative_2020}. Similarity-based self-supervised learning has shown great promise \cite{banville_uncovering_2021} and remains an underexploited avenue for EEG representation learning, especially for tasks extending beyond state decoding to long-term outcome prognosis. The fuller temporal trajectory of the transition, including slower and more intricate dynamics, is also largely unexploited, yet the path an individual brain takes through that transition, not only the depth it reaches, may carry relevant physiological information about its state.

Here, we develop ReMAP (Relative-positioning EEG Manifold for Anesthesia Prognosis), a representation-learning framework that turns raw intraoperative EEG into a 2D representation for perioperative brain-state monitoring and postoperative prognosis. We learn a similarity-based self-supervised representation suited to operating-room EEG (two frontal electrodes, raw waveforms, no labels) that places each EEG within a low-dimensional space. In this space, depth becomes one readable direction, while the geometry of the trajectory also separates how brain activity moves during anesthesia, whether by a rapid deepening, a passage through burst suppression, or a stable course. We evaluate this framework across two cohorts and two acquisition systems, for a total of over 1,000 patients, with four contributions. 

First, an interpretable representation in which anesthetic state occupies a space rather than a point, separating distinct courses that a depth of anesthesia index may consider similar. Second, evidence that the representation captures anesthetic depth label-efficiently and remains competitive with larger foundation models in the sparse-montage setting. Third, we show that age has its own gradient within the space, a correlate of brain ageing that the representation organizes without supervision. Fourth, in a secondary analysis of a prospectively enrolled cohort previously characterized with steady-state features, we propose that the early-phase trajectory through this space provides prognostic separation for 30-month cognitive and mortality outcomes.
\section{Results}
\subsection{Study design}
We analyzed high-frequency EEG-monitor recordings from the open-access VitalDB dataset (N=861) and the in-house LaribDB cohort (N=178). Patients were included based on anesthetic maintenance strategy (propofol and remifentanil), along with a high-resolution EEG available. Only VitalDB data were used for training our model. Patients were split at the patient level into training, validation, and test sets (60\%/20\%/20\%). Mean record duration was 4.37 (± 1.95) hours in the combined train/validation sets and 4.42 (± 1.95) in the test set. Mean age was 58.5 (± 14.9) and 59.3 (± 15.1), with 40.4 \% and 38.7 \% women, respectively. In LaribDB, four outcome groups were defined based on the occurrence of cardiovascular, cognitive decline or death events within 30 months after surgery. 

We compared feature learning strategies from raw EEG spanning classical power spectral density approaches (alpha/delta band power, burst suppression ratio), our custom self-supervised model along with its supervised backbone, and state-of-the-art pretrained foundation models (BENDR \cite{kostas_bendr_2021}, LUNA \cite{doner_luna_2025}, CBraMod \cite{wang_cbramod_2025} and REVE \cite{ouahidi_reve_2025}). Considering its larger size, only the VitalDB dataset was used for pretraining. 
\subsection{Pretraining benchmark}
First, we evaluate how accurately each representation learned through various pretraining methods can be used to predict the depth-of-anesthesia index BIS. On the VitalDB test set (N=173 patients), the SSL model predicted the BIS most accurately of all approaches (Fig. 1A-D), with a mean absolute error (MAE) of 3.2 (± 2.7), a negligible bias 0.1± 4.0, and a coefficient of determination R2 of 0.82. It outperformed the supervised model (MAE  4.6± 4.2, and R2 0.60), the EEG foundation models CBraMod (MAE 5.0 ± 4.6, R2 0.55), REVE (MAE 5.6 ± 5.2, R2 0.41), LUNA (MAE 5.4± 5.2 and R2 0.44) and BENDR (MAE 6.6± 6.0, R2 0.24), and classical PSD (MAE 5.85 ± 5.8, R2 0.32). Predicted versus true plots revealed differences depending on depth of anesthesia regimes. Whilst performing well in low-BIS regimes, PSD features fail to predict BIS in awake states accurately with a concordance correlation coefficient (CCC) of 0.50 (intercept 27.7, slope 0.33). Whilst trained on different datasets and a larger number of electrodes, foundation models tend to predict the general tendency with CCC scores ranging from 0.39-0.7 (intercept 32.4 to 19.9). ReMAP reached a CCC score of 0.90 (intercept 7.2, slope 0.83).

\begin{figure}
  \centering
  \includegraphics[width=0.9\textwidth]{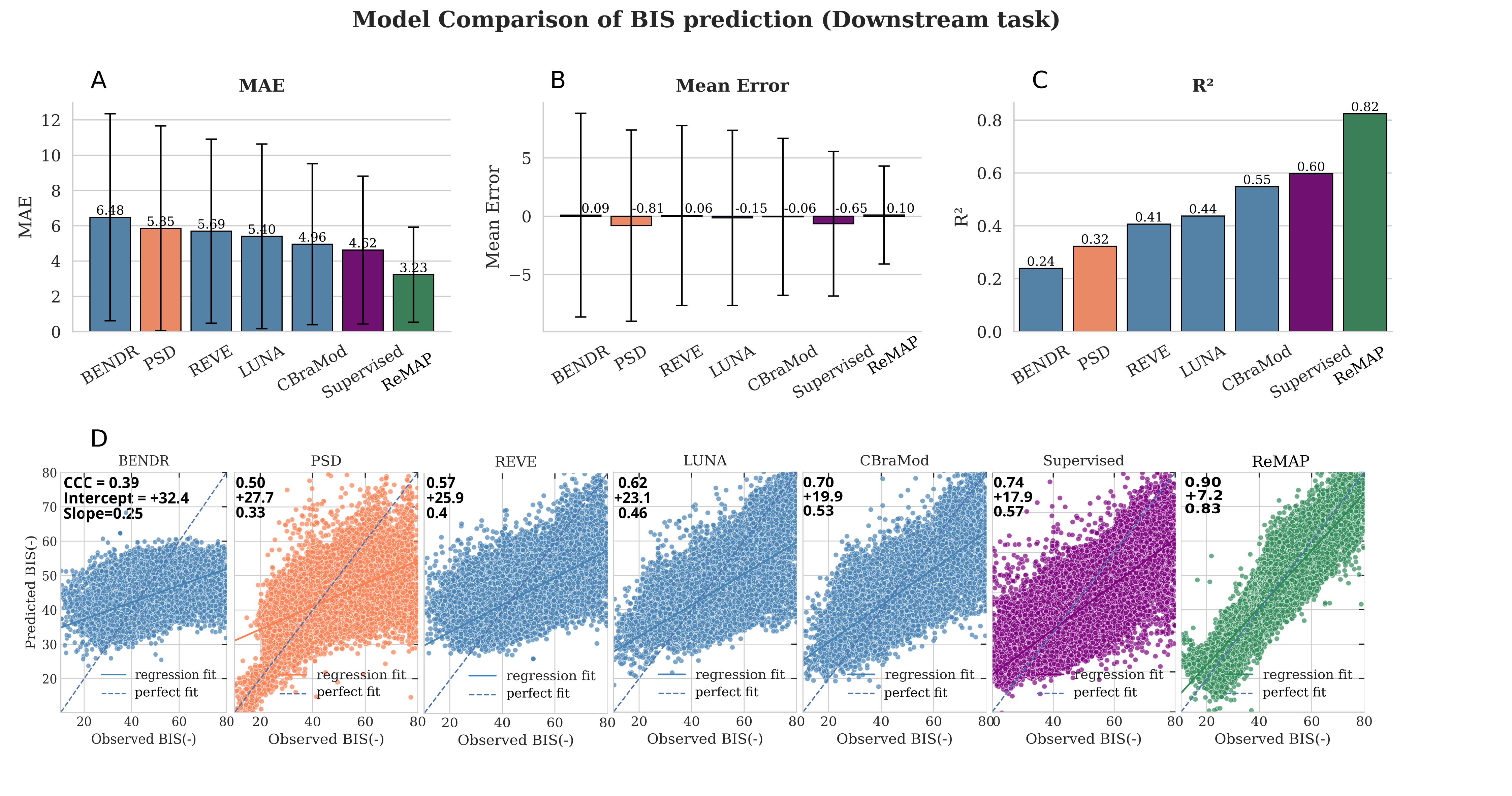}
\caption{Depth of anesthesia prediction on VitalDB test dataset ($n_{\text{patients}} = 173$, $n_{\text{windows}} = 84{,}444$). Foundation models are represented in blue, hand-engineered spectral features in orange, supervised in purple and ReMAP (Proposed) in green. \textbf{(A)} Mean absolute error per model. \textbf{(B)} Mean error (bias) per model. \textbf{(C)} $R^2$ reconstruction score per model. \textbf{(D)} Predicted vs observed BIS plots per model. Dashed line represents the perfect fit 1--1 line, and the solid line the obtained linear fit on models' predictions.}
  \label{fig:fig1}
\end{figure}

\subsection{Embedded physiological information}
\begin{figure}
  \centering
  \includegraphics[width=0.9\textwidth]{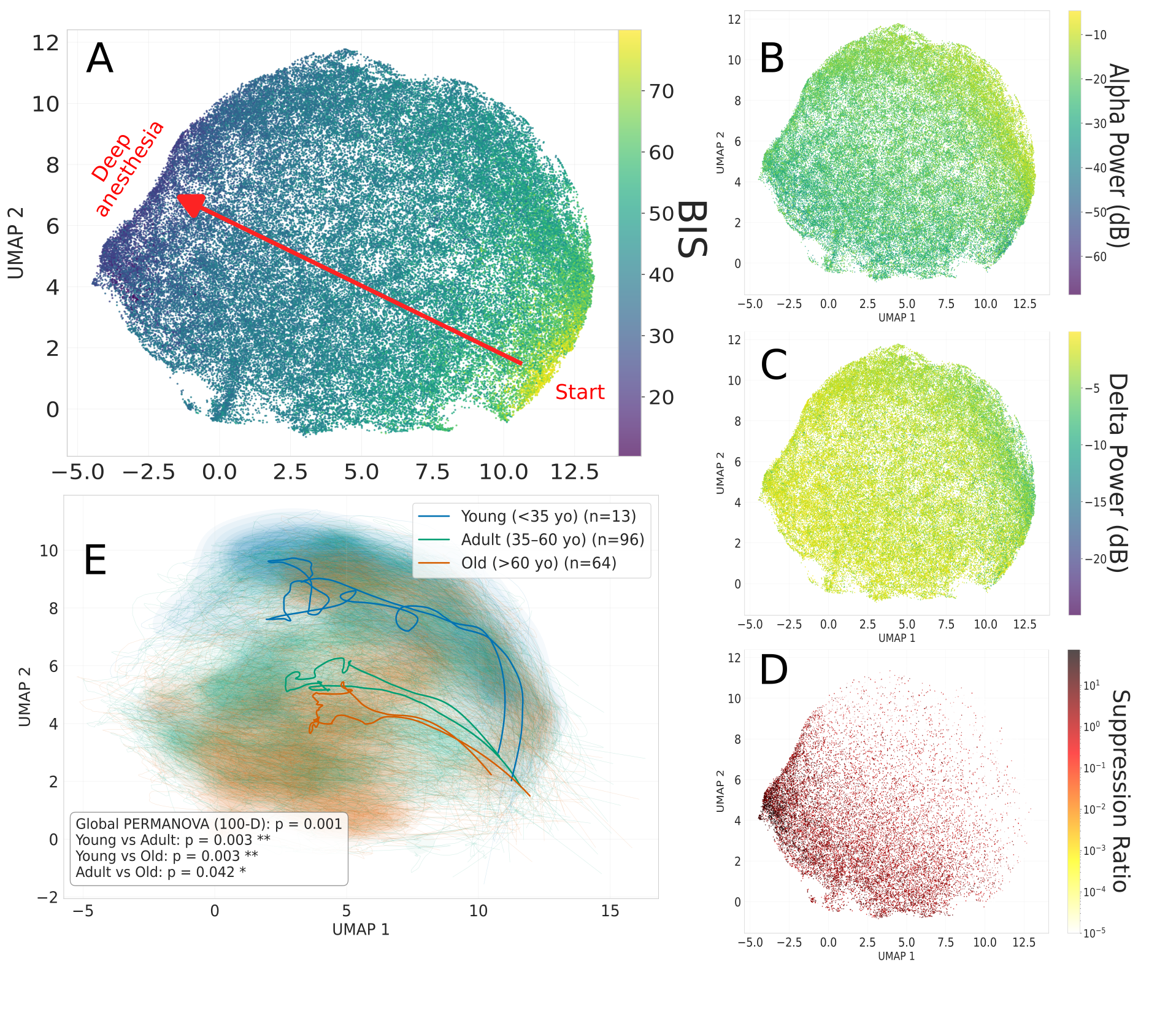}
  \caption{Embedding representation in 2D using UMAP dimensionality reduction on VitalDB test dataset ($n_{\text{patients}} = 173$, $n_{\text{windows}} = 84{,}444$). 2D projected embeddings colored by \textbf{(A)} Projected embeddings colored with the bispectral index. The red arrow indicates the direction of the drug injection and the passage from the beginning of the recordings -- awake -- to the anesthetized state. \textbf{(B)} Embeddings colored with the alpha band power (8--13~Hz). \textbf{(C)} Embeddings colored with the delta band power (0.5--4~Hz). \textbf{(D)} Burst suppression ratio. \textbf{(E)} UMAP trajectories of EEG dynamics across age groups. Individual patient trajectories (thin lines, $n = 13\,/\,96\,/\,64$ for Young\,/\,Middle-aged\,/\,Older) and group-wise median trajectories (thick lines, smoothed with a Savitzky--Golay filter) are projected onto the 2D UMAP embedding. Shaded background regions show the 2D kernel density estimate of each group's trajectory points, rendered as nine percentile bands (10th--90th, in 10\% increments) with darker shading indicating higher point density. Colors denote age group: Young ($\leq 35$~yo, blue), Middle-aged (35--60~yo, green), Older ($>60$~yo, orange). PERMANOVA on the patient $\times$ patient mean pointwise Euclidean distance matrix revealed a significant effect of age group on trajectory shape (pseudo-$F = 4.94$, $p = 0.001$); $p$-values are reported in the lower-left inset.}
  \label{fig:fig2}
\end{figure}

To investigate the structure of the learned representation using our SSL method, we project the embeddings into 2D using UMAP \cite{mcinnes_umap_2020}, and color them with various physiological properties. Quantitative statistical analysis is computed in the original 100-D latent space of the embedding. The corresponding embeddings for concurrent foundation models are shown in the supplementary figure S1. Our proposed representations exhibit a clear gradient along depth of anesthesia (BIS), with two distinct regions at BIS extremes (Fig.~\ref{fig:fig2}A). The BIS and alpha-power patterns were only weakly aligned (cosine $= 0.37$, 95\% CI [$+0.26$, $+0.47$]) (Fig.~\ref{fig:fig2}B--C), indicating largely complementary encoding, whereas BIS and suppression ratio were strongly and oppositely aligned (cosine $= -0.79$, 95\% CI [$-0.83$, $-0.74$]; ${\sim}142^\circ$) (Fig.~\ref{fig:fig2}D), consistent with the two variables encoding a shared, anti-correlated axis of anesthetic depth.
We next examine the age, showing median per-patient trajectories for three age groups: young ($\leq 35$~y, $n=13$), middle-aged (35--60~y, $n=96$), and older ($>60$~y, $n=64$) as shown in Fig.~\ref{fig:fig2}E. Age gradient was not aligned with depth-of-anesthesia gradient (cosine $= 0.09$, 95\% CI [$-0.12$, 0.32]). Trajectories progressed in an orderly fashion across groups, showing the learned manifold's capacity to discriminate among patients by age. We quantified this on the native manifold (100D), UMAP serving only for visualization. We found that the age group significantly affected trajectory shape (pseudo-$F = 4.94$, $p < 0.001$), with no significant heterogeneity of dispersion across age groups (global $F = 2.27$, $p = 0.134$). Pairwise comparisons were significant for Young vs Older ($p = 0.003$), Middle-aged vs Older ($p = 0.042$), and Young vs Middle-aged ($p = 0.003$).

\subsection{Validation on a dataset with outcomes}
\begin{figure}
  \centering
  \includegraphics[width=0.9\textwidth]{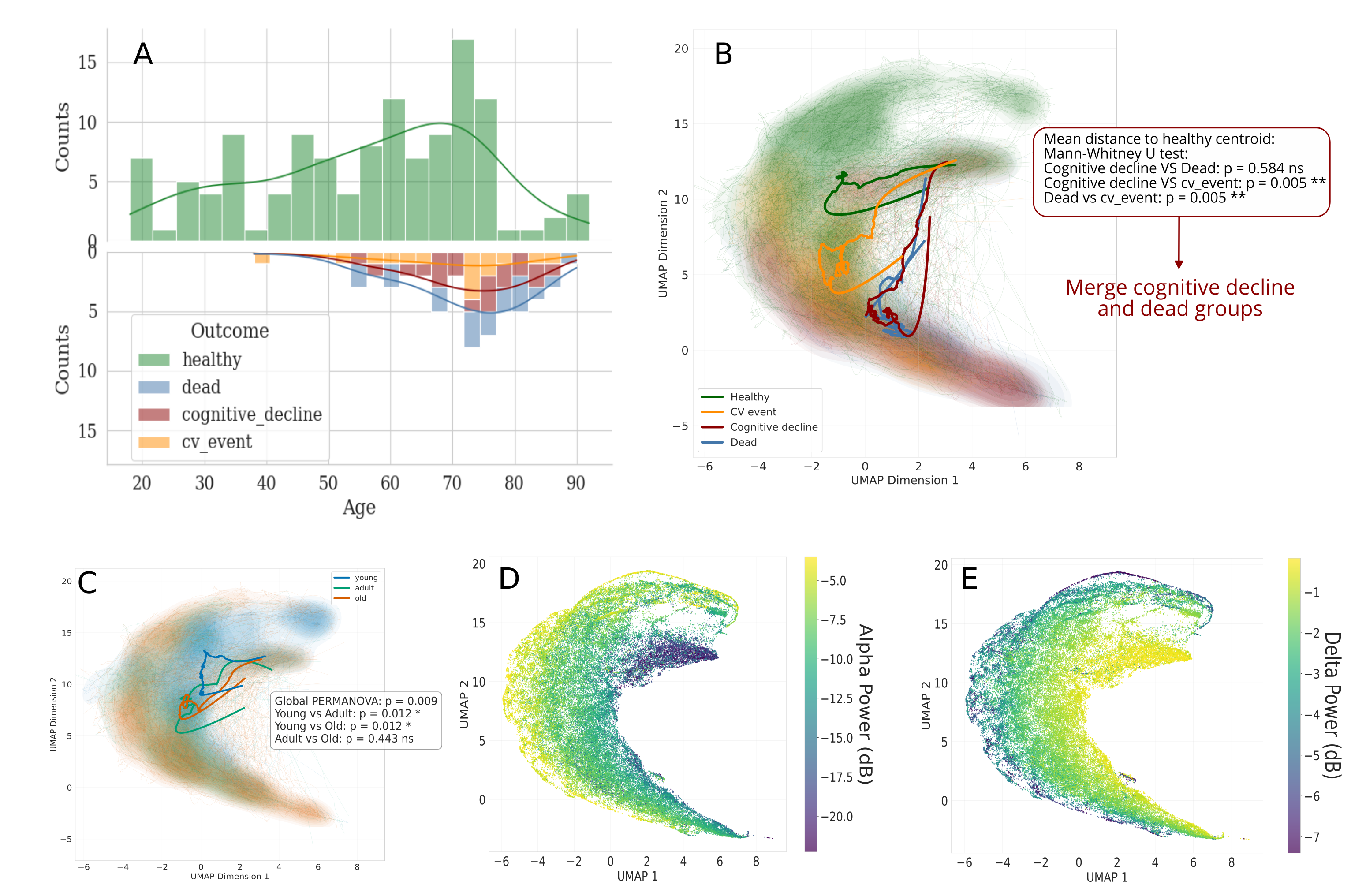}
 \caption{Model generalization to a clinical dataset with outcomes ($n_{\text{patients}} = 178$ shown; $n = 171$ after quality control for statistical tests, $n_{\text{windows}} = 58{,}974$). 2D projected embeddings colored by \textbf{(A)} Age distribution per outcome group. \textbf{(B)} Trajectories per outcome group. Shaded background regions show the 2D kernel density estimate of each group's trajectory points, rendered as nine percentile bands (10th--90th, in 10\% increments) with darker shading indicating higher point density. \textbf{(C)} Trajectories per age group. Shaded background regions show the 2D kernel density estimate of each group's trajectory points, rendered as nine percentile bands (10th--90th, in 10\% increments) with darker shading indicating higher point density. \textbf{(D)} Embeddings colored with alpha band power (8--13~Hz). \textbf{(E)} Embeddings colored with delta band power (0.5--4~Hz).}
  \label{fig:fig3}

\end{figure}

We then applied the VitalDB pretrained model, without fine-tuning, to the LaribDB dataset, for which 30-month outcome states were available: healthy (n=124, 69\%), cognitive decline (n=20, 11\%), death (n=19, 11\%), and cardiovascular event (n=15, 8\%) ; after quality control (Methods), 171 patients were retained for all outcome analyses. Outcome distribution against age is shown in Fig. \ref{fig:fig3}A, and the transferred embeddings in Fig. \ref{fig:fig3}B. In the embedding, cognitive decline and death trajectories occupied an overlapping region; therefore, we computed the mean distance of individual trajectories to the healthy group centroid in the original embedding space. It was significantly greater for cognitive decline and dead patients than for cardiovascular event patients (Mann-Whitney U, p=0.005 for both pairwise comparisons), while cognitive decline and dead patients did not differ from each other (p=0.584), consistent with a-priori grouping of these two outcomes. 

PERMANOVA in the original embedding space did not separate the groups. PERMDISP showed that within-group dispersion was likewise comparable (global F=2.12, p=0.123; all pairwise $p_{\text{holm}}$ $\geq$ 0.38), so the null result is not attributable to unequal dispersion. By contrast, the same test recovered age-related structure in this cohort: young vs middle-aged and young vs older, both significant (p=0.012); middle-aged and older patients were not (Fig. \ref{fig:fig3}C). Dispersion was not significantly heterogeneous across age groups (PERMDISP p=0.075).  The embeddings colored by alpha/delta power are displayed in Figure \ref{fig:fig3}D-E, depth of anesthesia was not available in this dataset.  

\begin{figure}
  \centering
  \includegraphics[width=0.9\textwidth]{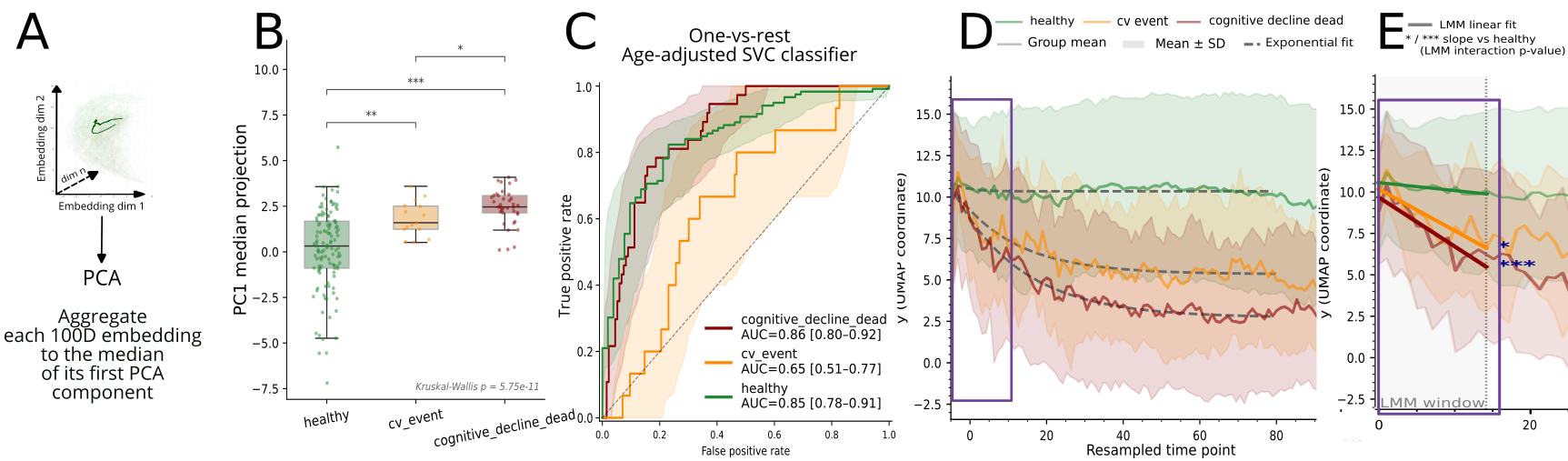}
  \caption{(A-C) Aggregated variable definition for outcome group classification (n = 171 after quality control). (D-E) Initial trajectory modelling. (A). Aggregation strategy of the 100-D trajectory into a 1-D vector based on PCA first component median (B) Aggregated feature distribution among outcome groups; asterisks denote statistical significance assessed by the two-sided Mann-Whitney U test (*p < 0.05, **p < 0.01, ***p < 0.001). (C) Receiver operating characteristic curves per outcome group; shaded bands represent 95\% bootstrap confidence intervals estimated over 1,000 resampling iterations. (D) Mean reconstructed trajectories over time per outcome group, obtained by resampling; shaded regions indicate the interquartile range (25th-75th percentile). Thick solid lines over the first 15 time points represent the group-level trajectories estimated by the linear mixed model (LMM); dashed lines show the fitted exponential decay model over the full observation window (E) Early induction dynamics (first 15\% of time points) of UMAP coordinate y by outcome group, with group-specific linear mixed-model fits. Asterisks at the right edge of the LMM window denote a statistically significant difference in slope relative to the healthy group (* p < 0.05; *** p < 0.001), as estimated from the LMM interaction term (time × outcome group).}
  \label{fig:fig4}
\end{figure}

Because PERMANOVA was underpowered, we tested whether condensed information from individual trajectories carried outcome information. To reduce each patient's trajectory to a single scalar, we used the per-patient median of the [PC1 / y] coordinate -- a robust summary chosen a priori, without reference to outcome labels (Figure \ref{fig:fig4}A). A Kruskal-Wallis test among the three outcome groups -- Figure \ref{fig:fig4}B -- was found significant, and a pairwise Mann-Whitney U test also discriminated the cv\_event and healthy groups ($p < 0.0001$), cognitive decline\_death vs healthy groups ($p < 0.0001$) and the cv\_event vs cognitive decline or dead groups ($p < 0.05$). After adjusting this metric for age, it was found predictive of the outcome group with an AUROC of 0.86 [0.80--0.92] for the cognitive decline or dead group, 0.65 [0.501--0.775] for the cv\_event group and of 0.85 [0.78--0.91] for the healthy group. The associated ROC curves are available in Figure \ref{fig:fig4}C. Confidence intervals for per-class AUC and ROC curves were estimated by bootstrap resampling of the pooled out-of-fold predictions \cite{efron_bootstrap_1985} (1,000 resamples, 95\% percentile intervals). Model performance was estimated using stratified nested cross-validation, with an outer 5-fold loop providing unbiased performance estimates and an inner 3-fold loop performing grid search over the SVM regularization parameter $C \in \{0.01, 0.1, 1, 10, 100\}$, optimizing balanced accuracy to account for class imbalance.

The fitted exponential decay model closely tracked the group-median EEG trajectories in the UMAP latent space for all three outcome groups (Figure \ref{fig:fig4}D) during the initial decrease phase. The model captures the initial decrease of the y-coordinate common to all groups, reflecting acute EEG changes during anesthesia. To assess the divergence of slopes among groups, and since the exponential model can be modelled linearly in the initial region, a linear mixed model was fitted to the y-coordinate of the EEG latent trajectory, with time point, outcome group, and their interaction as fixed effects, and per-patient random intercepts and slopes (Figure \ref{fig:fig4}E and Supplementary Table S5). At baseline, the reference group (healthy) showed an intercept of 10.74 (95\% CI [10.17, 11.31], $p < 0.001$), with no significant baseline differences for the cv\_event ($\beta = -0.41$, 95\% CI [$-2.14$, 1.32], $p = 0.644$) or cognitive decline/death groups ($\beta = -0.92$, 95\% CI [$-2.08$, 0.25], $p = 0.122$). The main effect of time was not significant in the reference group ($\beta = -0.050$, 95\% CI [$-0.111$, 0.012], $p = 0.115$). However, both clinical groups showed significantly steeper declines over time compared to the reference: the cognitive decline/death group declined an additional $-0.26$ units per time point (95\% CI [$-0.38$, $-0.13$], $p < 0.001$), a divergence that remained significant across 20 independent UMAP re-embeddings (Supplementary Table S7). The cv\_event group showed an intermediate decline in the reported embedding ($\beta = -0.21$, 95\% CI [$-0.40$, $-0.02$], $p = 0.028$), although it did not replicate across re-embeddings and is not interpreted as robust. These results indicate that while all groups share a similar EEG state prior to anesthesia, cognitive decline or death is associated with a reproducible, progressive divergence in the latent EEG space over the course of anesthesia.

\begin{table}[htbp]
\centering

\label{tab:encoder-performance}
\setlength{\tabcolsep}{4pt}
\renewcommand{\arraystretch}{1.3}
\resizebox{\textwidth}{!}{%
\begin{tabular}{lccccccc}
\toprule
\textbf{Encoder} & \textbf{Macro AUROC} & \textbf{AUC\textsubscript{cog\_decline\_dead}} & \textbf{AUC\textsubscript{cv\_event}} & \textbf{AUC\textsubscript{healthy}} & \textbf{F1\textsubscript{cog\_decline\_dead}} & \textbf{F1\textsubscript{cv\_event}} & \textbf{F1\textsubscript{healthy}} \\
\midrule
ReMAP
  & $0.786 \pm 0.032$
  & \shortstack{$0.862$ \\ {\footnotesize $[0.797\text{--}0.915]$}}
  & \shortstack{$0.645$ \\ {\footnotesize $[0.501\text{--}0.775]$}}
  & \shortstack{$0.847$ \\ {\footnotesize $[0.781\text{--}0.908]$}}
  & \shortstack{$0.625$ \\ {\footnotesize $[0.489\text{--}0.742]$}}
  & \shortstack{$0.222$ \\ {\footnotesize $[0.091\text{--}0.342]$}}
  & \shortstack{$0.695$ \\ {\footnotesize $[0.618\text{--}0.765]$}} \\
\midrule
PSD
  & $0.762 \pm 0.038$
  & \shortstack{$0.863$ \\ {\footnotesize $[0.796\text{--}0.920]$}}
  & \shortstack{$0.555$ \\ {\footnotesize $[0.423\text{--}0.672]$}}
  & \shortstack{$0.864$ \\ {\footnotesize $[0.807\text{--}0.919]$}}
  & \shortstack{$0.634$ \\ {\footnotesize $[0.512\text{--}0.750]$}}
  & \shortstack{$0.163$ \\ {\footnotesize $[0.037\text{--}0.285]$}}
  & \shortstack{$0.763$ \\ {\footnotesize $[0.689\text{--}0.823]$}} \\
\midrule
CBraMod
  & $0.767 \pm 0.029$
  & \shortstack{$0.818$ \\ {\footnotesize $[0.741\text{--}0.883]$}}
  & \shortstack{$0.643$ \\ {\footnotesize $[0.513\text{--}0.775]$}}
  & \shortstack{$0.835$ \\ {\footnotesize $[0.771\text{--}0.897]$}}
  & \shortstack{$0.511$ \\ {\footnotesize $[0.374\text{--}0.639]$}}
  & \shortstack{$0.120$ \\ {\footnotesize $[0.000\text{--}0.241]$}}
  & \shortstack{$0.764$ \\ {\footnotesize $[0.696\text{--}0.825]$}} \\
\midrule
REVE
  & $0.694 \pm 0.035$
  & \shortstack{$0.766$ \\ {\footnotesize $[0.686\text{--}0.844]$}}
  & \shortstack{$0.576$ \\ {\footnotesize $[0.427\text{--}0.729]$}}
  & \shortstack{$0.756$ \\ {\footnotesize $[0.684\text{--}0.834]$}}
  & \shortstack{$0.465$ \\ {\footnotesize $[0.324\text{--}0.600]$}}
  & \shortstack{$0.167$ \\ {\footnotesize $[0.038\text{--}0.293]$}}
  & \shortstack{$0.699$ \\ {\footnotesize $[0.624\text{--}0.767]$}} \\
\midrule
Supervised
  & $0.691 \pm 0.047$
  & \shortstack{$0.797$ \\ {\footnotesize $[0.724\text{--}0.865]$}}
  & \shortstack{$0.512$ \\ {\footnotesize $[0.363\text{--}0.666]$}}
  & \shortstack{$0.799$ \\ {\footnotesize $[0.726\text{--}0.870]$}}
  & \shortstack{$0.527$ \\ {\footnotesize $[0.400\text{--}0.652]$}}
  & \shortstack{$0.109$ \\ {\footnotesize $[0.000\text{--}0.222]$}}
  & \shortstack{$0.709$ \\ {\footnotesize $[0.636\text{--}0.777]$}} \\
\midrule
LUNA
  & $0.690 \pm 0.058$
  & \shortstack{$0.720$ \\ {\footnotesize $[0.633\text{--}0.805]$}}
  & \shortstack{$0.587$ \\ {\footnotesize $[0.441\text{--}0.730]$}}
  & \shortstack{$0.745$ \\ {\footnotesize $[0.667\text{--}0.823]$}}
  & \shortstack{$0.337$ \\ {\footnotesize $[0.217\text{--}0.473]$}}
  & \shortstack{$0.136$ \\ {\footnotesize $[0.000\text{--}0.279]$}}
  & \shortstack{$0.676$ \\ {\footnotesize $[0.602\text{--}0.744]$}} \\
\midrule
BENDR
  & $0.685 \pm 0.063$
  & \shortstack{$0.756$ \\ {\footnotesize $[0.674\text{--}0.833]$}}
  & \shortstack{$0.528$ \\ {\footnotesize $[0.388\text{--}0.672]$}}
  & \shortstack{$0.749$ \\ {\footnotesize $[0.672\text{--}0.824]$}}
  & \shortstack{$0.466$ \\ {\footnotesize $[0.333\text{--}0.577]$}}
  & \shortstack{$0.000$ \\ {\footnotesize $[0.000\text{--}0.000]$}}
  & \shortstack{$0.702$ \\ {\footnotesize $[0.632\text{--}0.769]$}} \\
\midrule
Age-only (baseline)
  & $0.711 \pm 0.074$
  & \shortstack{$0.757$ \\ {\footnotesize $[0.674\text{--}0.832]$}}
  & \shortstack{$0.572$ \\ {\footnotesize $[0.428\text{--}0.705]$}}
  & \shortstack{$0.755$ \\ {\footnotesize $[0.678\text{--}0.826]$}}
  & \shortstack{$0.489$ \\ {\footnotesize $[0.354\text{--}0.609]$}}
  & \shortstack{$0.093$ \\ {\footnotesize $[0.000\text{--}0.218]$}}
  & \shortstack{$0.706$ \\ {\footnotesize $[0.635\text{--}0.777]$}} \\
\bottomrule
\end{tabular}%

}
\caption{Models classification performance across outcome categories and competing methods controlled by age. The original embeddings information is aggregated using the median on PCA first axis before controlling against age. Area under the receiver operating characteristic curve (AUC, with 95\% bootstrap confidence intervals in brackets) are reported for three binary classification tasks: cognitive decline or death, healthy ageing, and cardiovascular event. Methods compared include our proposed model (ReMAP), the deep learning supervised baseline, EEG foundation models (CBraMod, REVE, BENDR and LUNA), and a classical power spectral density feature baseline (PSD). Higher values indicate better discriminative performance.}
\label{tab:tab1}
\end{table}

We also benchmarked our model against previous competing approaches (Table \ref{tab:tab1}). All predictions are controlled by age and downsampled to a single feature per patient. Our method achieves a macro AUROC ($0.786 \pm 0.032$), numerically ahead of the spectral baseline (PSD, 0.762) and the pretrained EEG foundation models (CBraMod, 0.767; REVE, LUNA, and BENDR, all $\leq 0.694$). On the two classes most strongly tied to the underlying signal -- cognitive decline/death and healthy -- our method is among the best performing, with AUCs of 0.86 and 0.85 and F1 scores of 0.62 and 0.69. Performance on the cardiovascular event class is markedly lower for every method (AUCs of 0.51--0.64, F1 scores of 0.00--0.19), which is expected given its weaker association with EEG data; our method nonetheless handles it best (class-2 F1 of 0.222, versus 0.000 for BENDR). Also, three of four foundation models do not beat chronological age alone in this task. Overall, these results show that our approach is competitive on the classes that are genuinely recoverable from the signal, while the limited performance on the cardiovascular event class reflects an inherent limitation of the task rather than of the method.

\FloatBarrier
\section{Discussion}

General anesthesia offers a rare opportunity to observe human brain function during a controlled physiological perturbation. In the operating room, this observation is usually reduced to estimating whether anesthesia is too light, appropriate, or too deep. Simplified frontal EEG has made it possible at bedside, and tools like BIS or PSI are routinely used for anesthetic titration. Yet these indices compress a complex, time-varying signal into a scalar value. This is useful but says little about how the brain reached that depth, how stable it is, or whether patients follow distinct courses. The relevant question is therefore not whether depth indices are useful, but whether part of the discarded geometry carries physiological or clinical meaning.

In this study, we investigated this question using similarity-based self-supervised learning on raw two-channel frontal EEG. Without labels or hand-engineered features, we construct a low-dimensional representation in which anesthetic depth can be recovered, while preserving aspects of the patient trajectory that may relate to age, brain vulnerability and postoperative outcome. We evaluate this approach across two cohorts and two acquisition systems, asking whether the learned geometry can support both depth estimation and a more individualized description of how the brain responds to anesthesia.

A first result is that the approach works on the sparse frontal montage found in operating rooms. Despite using only two frontal EEG channels, the SSL model recovered the BIS depth index more accurately than hand-engineered spectral features, a supervised model, and larger EEG foundation models, using only ~68k parameters against the 4M-157M of the foundation models. Similarity-based methods may arguably surpass reconstruction-based methods for this task \cite{banville_uncovering_2021}. One likely reason is that EEG has a low signal-to-noise ratio, for which reconstruction may emphasize waveform details that are not physiologically stable, while a similarity objective can learn brain-state changes from temporal continuity. It may also explain why REVE and LUNA, state-of-the-art on full multi-channel benchmarks, degrade more than models built for limited-montage clinical settings such as CBraMod and Chambon. This illustrates that matching models to downstream tasks, including montage compatibility and domain alignment of pretraining data, can dominate raw model scale \cite{kokate_channel_2026}, and complements the work already underway within the community to evaluate the performance of EEG foundation models in clinical tasks \cite{banville_neuralbench_2026, guetschel_s-jepa_nodate}. Because the depth target is itself a proprietary index (BIS), these results inherit its known limitations; the value of the representation therefore lies less in reproducing BIS than in the additional trajectory structure it exposes.

The learned representation was not only predictive of depth, but also appeared to organize physiological structure without supervision. In the VitalDB cohort, depth evolved in a clear direction, while age, the principal confounder of vulnerability, showed a separated trajectory across the manifold. Age group significantly affected trajectory shape, and the age gradient showed little measurable alignment with the depth axis, with a cosine similarity close to zero. This should not be seen as proof of independence between these axes, but rather that the representation captures age-related variation not reducible to depth alone. Similarly, the SSL representation captures physiological variables such as alpha or slow-delta band powers, the main oscillations under anesthetics targeting primarily GABA receptors such as propofol (extension to halogenated vapors remains to be tested)\cite{purdon_electroencephalogram_2013, akeju_effects_2014}, as well as burst-suppression. This connection bridges clinically grounded and traditional biomarkers \cite{cartailler_alpha_2019, guessous_intraoperative_2023} (alpha, slow-delta, burst-suppression), whose physiology is well understood through thalamocortical dynamics \cite{guay_clinical_2025}, with artificial intelligence, whose representations are often criticized as uninterpretable \cite{rudin_stop_2019}.

In the independent LaribDB cohort, this trajectory-level information was associated with 30-month cognitive and survival outcomes. Our pipeline is competitive with both traditional EEG biomarkers and domain foundation models, providing data-driven biomarkers complementary to age. The most interpretable evidence came from the latent trajectory during induction (the early phase), where patients who later developed cognitive decline or died, and to a lesser extent those with cardiovascular events, had a steeper decline of the y-coordinate over time, despite a comparable baseline response. This is consistent with previous work suggesting that induction is a critical window for capturing individual vulnerability \cite{cartailler_brain_2021}, as the brain’s early response to anaesthetic exposure may reveal differences in reserve or resilience.

The observation that cognitive-decline and death trajectories overlap is consistent with prior work coupling worsening cognition to mortality risk \cite{luck_mortality_2015}, and with the view that brain dysfunction is an early marker of, and tightly coupled to, systemic deterioration, with cognitive impairment and brain failure linked to frailty, multisystem physiological decline, and, in critical illness, the cascade of multi-organ failure \cite{lobo_cognitive_2025, chen_longitudinal_2023, wang_cognitive_2020, michelagnoli_organ_2013}.

The present work has several limitations. The outcome cohort is single-centre, observational, and modest in size. The sample size of cardiovascular event was small. For these reasons the model was transferred to LaribDB without fine-tuning. Classification was weakest for cardiovascular events, which is expected, since frontal EEG reflects cardiovascular pathophysiology only indirectly; this class is the natural target for complementary modalities such as the ECG, PPG, and high-frequency arterial pressure rather than EEG alone, although a dedicated pretext task for these pseudo-periodic signals, likely different from temporal similarity, would be needed to bring them into the same representation \cite{gopal_3kg_2021}. Finally, pooling cognitive decline and death into a single outcome carries a competing-risks structure as a cognitive-decline event requires surviving to the t+30 months MoCA assessment, whereas death is a competing event. Their overlapping trajectories and indistinguishable distance to the healthy centroid support the pooling, but future analyses should model the competing risks explicitly.

The next step is a prospective external validation in a larger, if possible, multi-centre cohort with multimodal fusion to better capture the cardiovascular risk. Beyond that, the trajectories invite a geometric approach, using tools from topology such as persistence diagrams \cite{bubenik_statistical_2012, goodman_persistent_2008} that may help parameterize the shape of a patient’s path quantitatively, toward an objective, topological characterization of the data manifold\cite{chazal_structure_2013}, possibly contributing to a universal, low-dimensional map of brain functions \cite{krumm_toward_2025}.

Together, these results suggest that the path a brain traces through anesthesia can be read as a shape rather than a single number, anesthetic depth is one axis of a low-dimensional space, while the geometry of the trajectory carries additional, physiologically meaningful structure. Learned from two frontal electrodes without labels, the representation recovers depth efficiently, separates age along its own direction, and, in an independent cohort, tracks 30-month cognitive and survival outcomes. These are associational, single-cohort findings, if confirmed prospectively, such trajectory geometry could provide an interpretable, label-efficient marker for perioperative risk stratification.

\section{Materials and Methods}
\subsection{Datasets and preprocessing}
The LaribDB included 178 patients from Lariboisière Hospital. The cohort characteristics are summarized in the supplementary material. Patients belonged to one of the four outcome groups: healthy, cognitive decline, cardiovascular event and death. These outcomes were defined 30 months after data acquisition. Cognitive decline was defined as a drop of more than one standard deviation between the t and t+30 months Montreal Cognitive Assessment (MoCA) test. EEG was monitored using the Sedline device from Masimo (f=500Hz, resampled to 100Hz). The VitalDB dataset included 861 patients and was downloaded using the provided vitaldb (version 1.4.11) library \cite{lee_vitaldb_2022}. The patients relevant to this study were selected based on the simultaneous presence of waveform 2-lead EEG, BIS, PPG, ECG, HR, arterial blood pressure and propofol injection signals. EEG was monitored with the Medtronic Vista sensor. Data were recorded at 1 Hz for low-frequency data and 128 Hz for high-frequency waveform data. Patients were maintained unconscious using propofol and remifentanil. EEG waveforms were filtered using a low-pass filter at 30 Hz. The dataset was split into train, validation, and test sets at the patient level, with a 60/20/20 ratio. EEG windows with mean BIS values outside of the [10, 80] range were excluded. The records were processed using the Braindecode \cite{aristimunha_braindecode_2025} Baseconcat datasets framework (version 1.5.0). Pretrained weights of EEG foundation models were downloaded from the corresponding artifacts repository on HuggingFace (huggingface-hub version 0.19.4). 
\subsection{Training pipeline}
\begin{figure}
  \centering
  \includegraphics[width=0.9\textwidth]{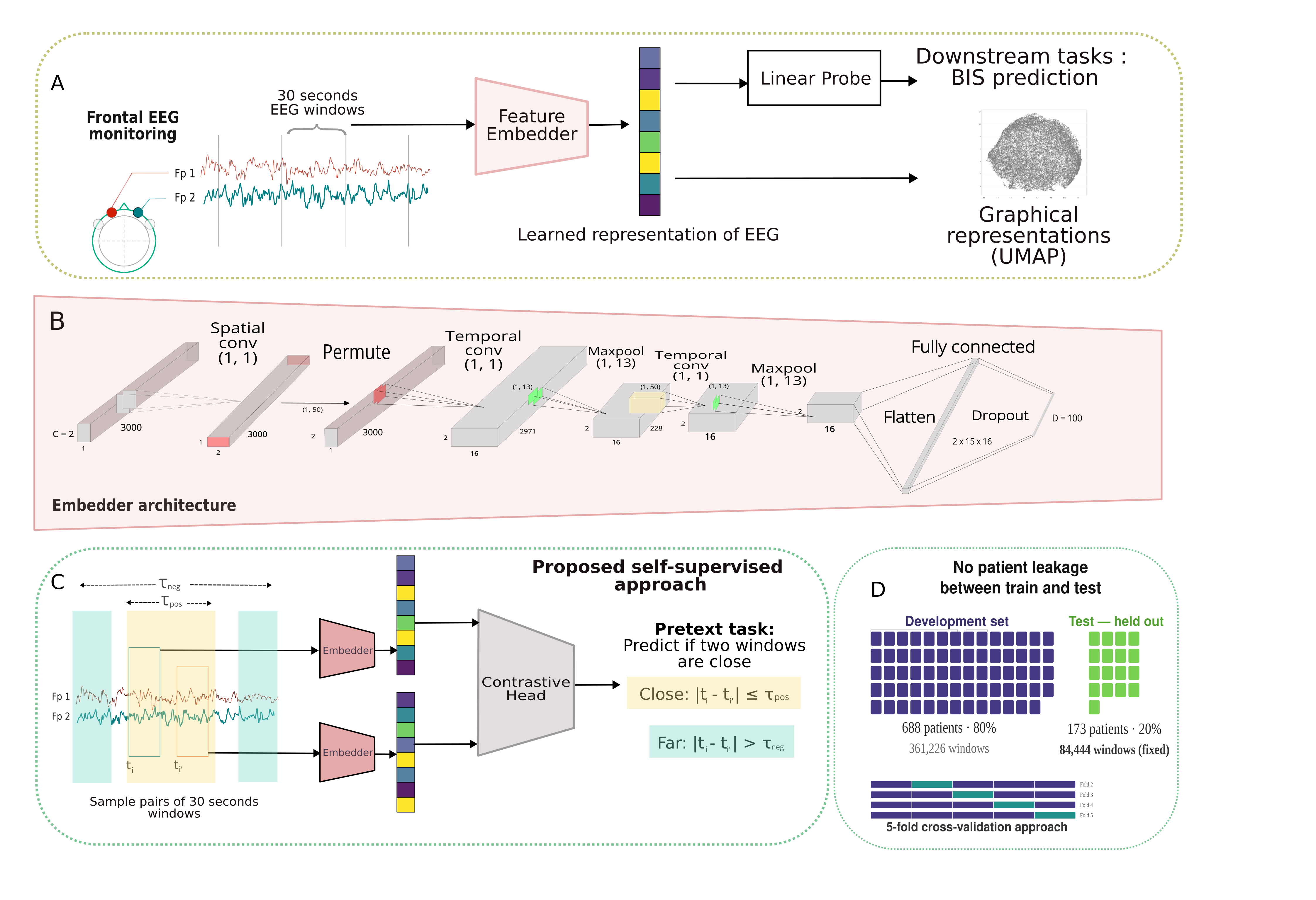}
  \caption{Depth of anesthesia prediction pipeline and benchmark. (A). General study design. (B). Proposed EEG embedder. (C). Pretraining computational framework and proposed self-supervised approach.  (D). Development and test set splitting strategy. }
  \label{fig:fig5}
\end{figure}

We present a model based on contrastive learning on the EEG. The proposed model is based on relative positioning, a self-supervised pretext task where the model learns to predict whether two EEG windows are close together or far apart in time \cite{mcinnes_umap_2020}. Temporal proximity is defined using the following definition: Given $\tau_{\text{pos}} \in \mathbb{N}$, which controls the duration of the positive context, and $\tau_{\text{neg}} \in \mathbb{N}$, which corresponds to the negative context around each window $X_i$, we sample $N$ labelled training pairs $\mathcal{T}$:
\begin{equation}
\mathcal{T} = \left\{ (t, t') \in \{M-L+1\}^2 \;\middle|\; |t - t'| \leq \tau_{\text{pos}} \;\text{ or }\; |t - t'| > \tau_{\text{neg}} \right\},
\end{equation}
\begin{equation}
y(t, t') =
\begin{cases}
1, & \text{if } |t_i - t_{i'}| \leq \tau_{\text{pos}} \\[4pt]
0, & \text{if } |t_i - t_{i'}| > \tau_{\text{neg}}
\end{cases} \; .
\end{equation}
In this setting, the data was split into 30-second consecutive segments. Same-recording window pairs were sampled, and their positive or negative label was generated according to their temporal separation. The pretraining loss is a binary logistic loss: a shared encoder corresponding to Figure \ref{fig:fig5}B embeds each window, and the element-wise absolute difference of the two embeddings is passed through a linear layer to predict the label.

\subsection{Downstream benchmarking and evaluation}

We benchmarked our approach against state-of-the-art methods spanning increasing model complexity. As a supervised baseline, the encoder CNN backbone was trained end-to-end with a linear regression head. We also evaluated recent EEG foundation models: BENDR, which combines a convolutional encoder with a transformer contextualizer for transfer learning across EEG tasks; REVE, a 69M-parameter transformer pretrained on approximately 60,000 hours of EEG from 25,000 subjects using 4D positional encodings to support arbitrary electrode configurations; and CBraMod, a smaller (~4M-parameter) model pretrained on the Temple University Hospital corpus; using a criss-cross spatial and temporal attention with asymmetric conditional positional encoding to handle arbitrary channel layouts. No baselines were fine-tuned and followed the same downstream protocol as for our own embeddings. 
Downstream tasks comprised BIS regression and, where available, outcome-group classification. Patients were divided into three age groups based on the age distribution of the VitalDB dataset: young ($\leq$ 35), middle-aged (35-60), and older ($>$ 60). For each model, the pretrained encoder was extracted and used to generate embeddings, which were passed to a shallow linear probe for task of interest. For each patient, the sequence of per-window embeddings was first z-scored globally using the mean and standard deviation computed across all windows from all patients. A patient-level quality-control step was then applied: the per-patient mean z-scored embedding was projected onto two principal components (PCA, fitted on the patient-level means), and patients whose projection exceeded ±3 SD on either PC1 or PC2 were excluded as outliers, independently of outcome (7 of 178 removed; 171 retained). A second PCA was then fitted on all z-scored windows from retained patients. Each patient's temporal window sequence was resampled to a fixed length of 100 equally-spaced steps to account for variable recording durations. The resampled trajectory was then projected onto the first principal component, and the median of these 100 PC1 values was taken as the scalar biomarker. Classification was performed using a linear SVC in a nested cross-validation scheme (5 outer folds, 3 inner folds for hyperparameter selection), with all preprocessing - including z-scoring and PCA fitting - performed exclusively on training data within each fold to prevent data leakage. 

Performances are always displayed for the unseen test set. As for the metrics used, models were evaluated based on classification and regression performance on downstream tasks of interest. Error and absolute error distributions, along with R2 scores, were used for regression tasks, while macro average area under the curve (AUC) and class-level F1 score were reported for classification.  Age was included as a covariate to control against, given its well-established association with cardiovascular risk and its known influence on EEG spectral features \cite{ayeb_using_2025, obert_influence_2021}, allowing the model to capture complementary demographic and neurophysiological information. While downstream performance quantifies the utility of the SSL-learned features, it does not reveal what structure they capture. To address this, we projected the 100-dimensional embeddings obtained from VitalDB to two dimensions using Uniform Manifold Approximation and Projection (UMAP)\cite{mcinnes_umap_2020}. This manifold learning technique for nonlinear dimensionality reduction was chosen for its ability to preserve the topological structure of high-dimensional data. Using this representation method, data and notably dataset changes are subject to implying notable changes in the obtained representations.
\subsection{Dynamics assessment}
Patient-level trajectories were constructed from UMAP embeddings resampled to $T=100$ time points. Trajectories were smoothed with a Savitzky--Golay filter (window $=51$, order $=3$) to reduce embedding noise while preserving local curvature. Because the second coordinate $y$ was the most informative, group-level median trajectories were fitted using an exponential resistor--capacitor decay model of the form $y(t) = A e^{-t/\tau} + C$, where $A$, $\tau$, and $C$ are amplitude, time constant, and vertical offset. Parameters were estimated separately for each outcome group by minimizing the mean squared error between the model and the group-median trajectory over the 100 time points, using L-BFGS-B optimization.

For the induction period, the exponential model linearizes for $t \ll 1$ to $y(t) \approx (A+C) - \frac{A}{\tau}\, t$ to the leading order, motivating a linear mixed-effect model (LMM, Python library \texttt{statsmodels}, version 0.14.6) analysis of the first 15 trajectory points of the induction period. Using the healthy group as reference, we fitted:
\begin{equation}
y_{ij} = \beta_0 + \beta_1 t_{ij} + \beta_2\, CV_i + \beta_3\, CD_i + \beta_4\, t_{ij} CV_i + \beta_5\, t_{ij} CD_i + u_{0i} + u_{1i} t_{ij} + \varepsilon_{ij},
\end{equation}
where $y_{ij}$ is the embedding coordinate for patient $i$ at time point $j$, $t_{ij}$ is the time, $CV_i$ and $CD_i$ indicate cardiovascular-event and cognitive-decline/death outcomes, and the healthy group corresponds to $CV_i = CD_i = 0$. $\beta_0$ is the intercept, $\beta_1$ the slope of $y$ over time in the healthy group, and $\beta_2$, $\beta_3$ the baseline offset of each group relative to healthy. The interaction coefficients $\beta_4$, $\beta_5$ quantify group-specific differences in early trajectory slope relative to healthy patients, while $u_{0i}$ and $u_{1i}$ model patient-specific random intercept and slope ($\varepsilon_{ij}$ is the residual error).

\subsection{Statistical and geometric analysis}
For each clinical variable, we asked whether changes in that variable followed a consistent direction in the 100D embedding space. This direction was estimated as an encoding pattern, defined as the unit-normalized covariance vector between each z-scored embedding dimension and the mean-centered variable. Alignment between variables was quantified as the cosine similarity between their encoding patterns ($0 =$ orthogonal, $1 =$ collinear), with 95\% confidence intervals estimated by patient-level bootstrap resampling over 2,000 iterations. Distance to the healthy centroid was computed as each patient's mean Euclidean distance to the healthy centroid trajectory (in the original 100D embeddings). Group differences in trajectories were assessed using the non-parametric permutational analysis of variance test (PERMANOVA) \cite{anderson_new_2001} on pairwise matrices. Additionally, PERMDISP was used to test whether effects reflected centroid shift rather than unequal dispersion. Both tests used \texttt{scikit-bio} 0.7.2 with 999 permutations, reporting pseudo-$F$ statistics and $p$-values. Group differences in continuous variables were tested using Mann--Whitney U or Kruskal--Wallis tests, as appropriate. All $p$-values were adjusted using Holm--Bonferroni correction \cite{holm_simple_1979}.

\section{Funding}
This research was funded by the French National Research Agency (ANR) via the PEPR Santé Numérique Diip Heart within the France 2030 programme, under the reference ANR-22-PESN-0018 and Anesth-IA, under the reference ANR-24-CE45-2602.

\section{Data availability}
VitalDB dataset: https://vitaldb.net/dataset/ The acquisition and release of the data was approved by the Institutional Review Board of Seoul National University Hospital (H-1408-101-605). The study was also registered at clinicaltrials.gov (NCT02914444). LaribDB dataset: All participants provided oral informed consent. Study was approved by the Committee for the Protection of Individuals (2019-A03206-51 and 2022-A02829-34). The data will be made available from authors upon reasonable requests. The readers can contact Jérôme Cartailler for more information.

\section{Code availability}
Custom code used for data analysis in this study will be deposited in a GitHub repository and archived on Zenodo with a permanent DOI upon publication.

\label{sec:headings}

%Bibliography
\bibliographystyle{unsrt}  
\bibliography{references}

\end{document}